\documentclass[10pt,twocolumn,letterpaper]{article}

\usepackage[algorithms]{wacv}      
\usepackage{wacv}

\usepackage{graphicx}
\usepackage{amsmath}
\usepackage{amssymb}
\usepackage{booktabs}
\usepackage[accsupp]{axessibility}

\usepackage{subcaption}
\usepackage{pifont}
\usepackage{multirow}
\usepackage{makecell}
\newcommand{\xmark}{\ding{55}}
\newcommand{\cmark}{\ding{51}}
\newcommand{\centered}[1]{\begin{tabular}{l} #1 \end{tabular}}

\usepackage[pagebackref,breaklinks,colorlinks]{hyperref}

\usepackage[capitalize]{cleveref}
\crefname{section}{Sec.}{Secs.}
\Crefname{section}{Section}{Sections}
\Crefname{table}{Table}{Tables}
\crefname{table}{Tab.}{Tabs.}

\def\wacvPaperID{1119} 
\def\confName{WACV}
\def\confYear{2024}

\begin{document}

\title{STEP - Towards Structured Scene-Text Spotting}

\newcommand{\titlespace}[0]{\quad}
\author{Sergi Garcia-Bordils$^{1,2}$
\titlespace
Dimosthenis Karatzas$^1$
\titlespace
Marçal Rusiñol$^2$\\
$^1$Computer Vision Center, UAB, Spain\\
$^2$AllRead MLT\\
{\tt\small \{sergi.garcia, dimos\}@cvc.uab.cat}
}
\maketitle

\begin{abstract}
    We introduce the structured scene-text spotting task, which requires a scene-text OCR system to spot text in the wild according to a query regular expression. Contrary to generic scene-text OCR, structured scene-text spotting seeks to dynamically condition both detection and recognition on user-provided regular expressions.
    To tackle this task, we propose the Structured TExt sPotter (STEP), a model that exploits the provided text structure to guide the OCR process. STEP is able to deal with regular expressions that contain spaces and it is not bound to detection at word-level granularity. Our approach enables accurate zero-shot structured text spotting in a wide variety of real-world reading scenarios and is solely trained on publicly available data.
    To demonstrate the effectiveness of our approach, we introduce a new challenging test dataset that contains several types of out-of-vocabulary structured text, reflecting important reading applications such as weight information, serial numbers, license plates etc.
    We demonstrate that STEP can provide specialized OCR performance on demand in all tested scenarios.
    The code and test dataset are released at \url{https://github.com/CVC-DAG/STEP}.
\end{abstract}

\section{Introduction}

A lot of textual content that appears in the world around us carries important semantic information, useful for numerous real-world applications. Examples include prices, dates, license plates, serial numbers, consumption readings on utility meters, URLs, telephone numbers, etc. Although scene text detection has advanced significantly over the past decade, current methods still fail to deal satisfactorily with out-of-vocabulary strings and text in dense configurations, which corresponds exactly to the type of cases of real-life interest.

In this work we propose a new model capable of extracting specific text in the wild on demand, as required by the end application. To do so, we exploit the fact that the sought after text has a specific structure to guide both the detection and recognition process through a query regular expression.

State of the art scene text recognition methods aim to recognize all text in the scene indiscriminately. Significant progress has been achieved in end-to-end scene text detection and recognition\cite{feng2019textdragon, liu2020abcnet, liao2020mask, liu2021abcnet, kim2022deer, peng2022spts, kittenplon2022towards, zhang2022text, huang2022swintextspotter} including in challenging scenarios such as curved text\cite{ch2017total, yuliang2017detecting}, text in video\cite{karatzas2013icdar, garcia2022read, reddy2020roadtext}, or multi-lingual settings\cite{nayef2017icdar2017, nayef2019icdar2019}.

\begin{figure}
    \centering
    \begin{tabular}{cc}
        \begin{subfigure}{0.42\columnwidth}
            \includegraphics[width=\columnwidth, trim={0 0 0 0.7cm},clip]{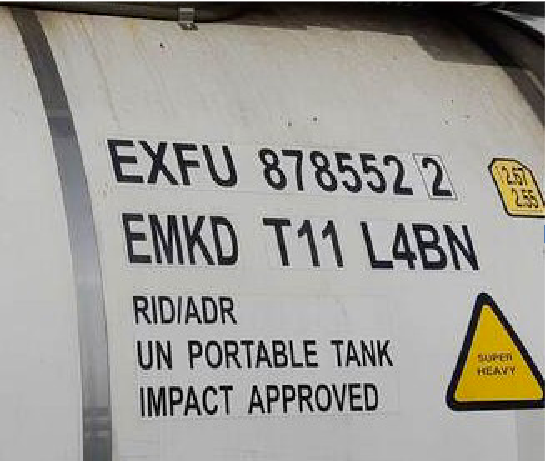}
            \caption{}
            \label{fig:intro_example_1}
        \end{subfigure}&
        \begin{subfigure}{0.42\columnwidth}
            \includegraphics[width=\columnwidth, trim={0 0 0 0.51cm},clip]{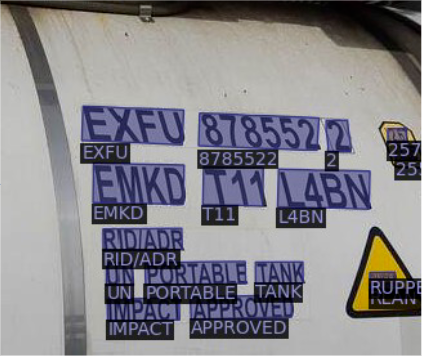}
            \caption{}
            \label{fig:intro_example_2}
        \end{subfigure}\\

        \begin{subfigure}{0.42\columnwidth}
            \includegraphics[width=\columnwidth, trim={0 0 0 0.41cm},clip]{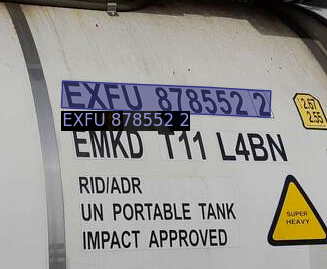}
            \caption{}
            \label{fig:intro_example_3}
        \end{subfigure}
        &
        \begin{subfigure}{0.42\columnwidth}
            \includegraphics[width=\columnwidth, trim={0 0 0 1.2cm},clip]{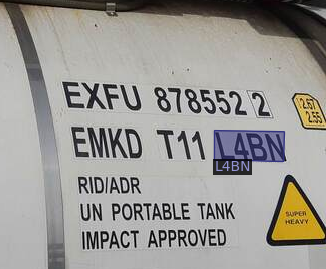}
            \caption{}
            \label{fig:intro_example_4}
        \end{subfigure}\\
    \end{tabular}

    \caption{We propose a model that dynamically conditions both scene text detection and recognition on user provided regular expressions. \ref{fig:intro_example_1} Image with different types of structured text. \ref{fig:intro_example_2} Results obtained with TESTR\cite{zhang2022text}. \ref{fig:intro_example_3} Result of the proposed method for regular expression ``\texttt{[A-Za-z]\{4\}\textbackslash s\textbackslash d\{6\}\textbackslash s\textbackslash d}''. \ref{fig:intro_example_4} Results of applying a different regex (``\texttt{[A-Za-z]\textbackslash d [A-Za-z]\{2\}}'').}
    \label{fig:intro_example}
\end{figure}

Current methods rely implicitly or explicitly on a language model, which might be acquired by the vocabulary of the training set (implicitly learnt)\cite{wan2020vocabulary, garcia2023out} or explicitly integrated in the recogniser \cite{fang2021read}.
Numerous recent evidence has demonstrated that such methods tend to over-rely on their language model~\cite{garcia2023out, wan2020vocabulary}, resulting in higher recognition error rates on out-of-vocabulary text, which is exactly the type of text which is most important for many real-world applications.
In addition, all methods perform detection at word-level granularity, while the information of interest might require combinations of various detected tokens. Recovering such information requires heuristic post-processing to combine detected word-level tokens and is prone to prior detection errors.
It is possible to train specific systems for extracting specific types of information\cite{gomez2018cutting}, but it is a cumbersome approach that requires application-specific annotated data, which is not easy to source.

\begin{figure}
   \centering
   \includegraphics[width=0.8\columnwidth]{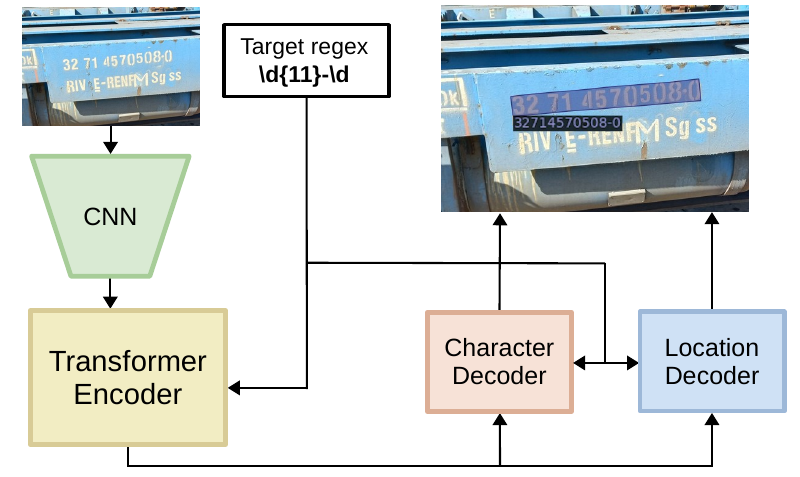}
   \caption{The STEP architecture is composed of a transformer\cite{vaswani2017attention} encoder and two character and location decoders, which are guided by the queried regex. The image serves as input to the CNN backbone, while the regex query is used as input by the encoder and decoders.}
   \label{fig:architecture_overview}
\end{figure}

We propose instead to train a single model that can dynamically condition both the detection and recognition stages on a user provided regular expression (regex). The text structure is provided on demand, and during testing the proposed model is employed in a zero-shot fashion with regular expressions never used during training. The model is capable of spotting only the relevant information in the scene, efficiently suppressing any other distracting text. The regex supported can contain spaces, therefore guiding the detection process from the very beginning to segment the scene text at the required granularity, not imposing any design restrictions to perform detection at the word level. 
Figure \ref{fig:architecture_overview} shows an overview of our proposed architecture, the Structured TExt sPotter (STEP).

We compare the proposed system with multiple state of the art methods. We demonstrate that state of the art methods are not capable of correctly detecting and recognising structured text, even if heuristic post-processing is applied on the recognition results. Furthermore, we put forward a new test set reflecting numerous real-life application scenarios where structured text is important and demonstrate the superiority of the proposed method in a zero-shot scenario, spotting regular expressions never seen during training.

The contributions of this paper can be summarized as follows:

\begin{itemize}
    \item We propose a new structured scene-text spotting task, where methods are expected to detect and recognise text in the wild that respects a dynamically provided query regular expression.    
    \item A challenging test dataset that contains several types of out-of-vocabulary structured text. Each image contains one or more instances of text that respect specific regular expressions. The dataset features also space-separated codes, which poses a particular challenge to generic OCR systems.
    \item The Structured TExt sPotter (STEP), a network where the detection and the recognition processes are guided with a queried regular expression. The model has been trained on generic, publicly available data, and it is not fine-tuned for any of the test cases.
    \item We perform comprehensive experimentation and ablation studies, and demonstrate that our approach outperforms state of the art scene-text baselines.
\end{itemize}

\begin{figure*}[h!]
    \centering
    \includegraphics[width=0.85\linewidth]{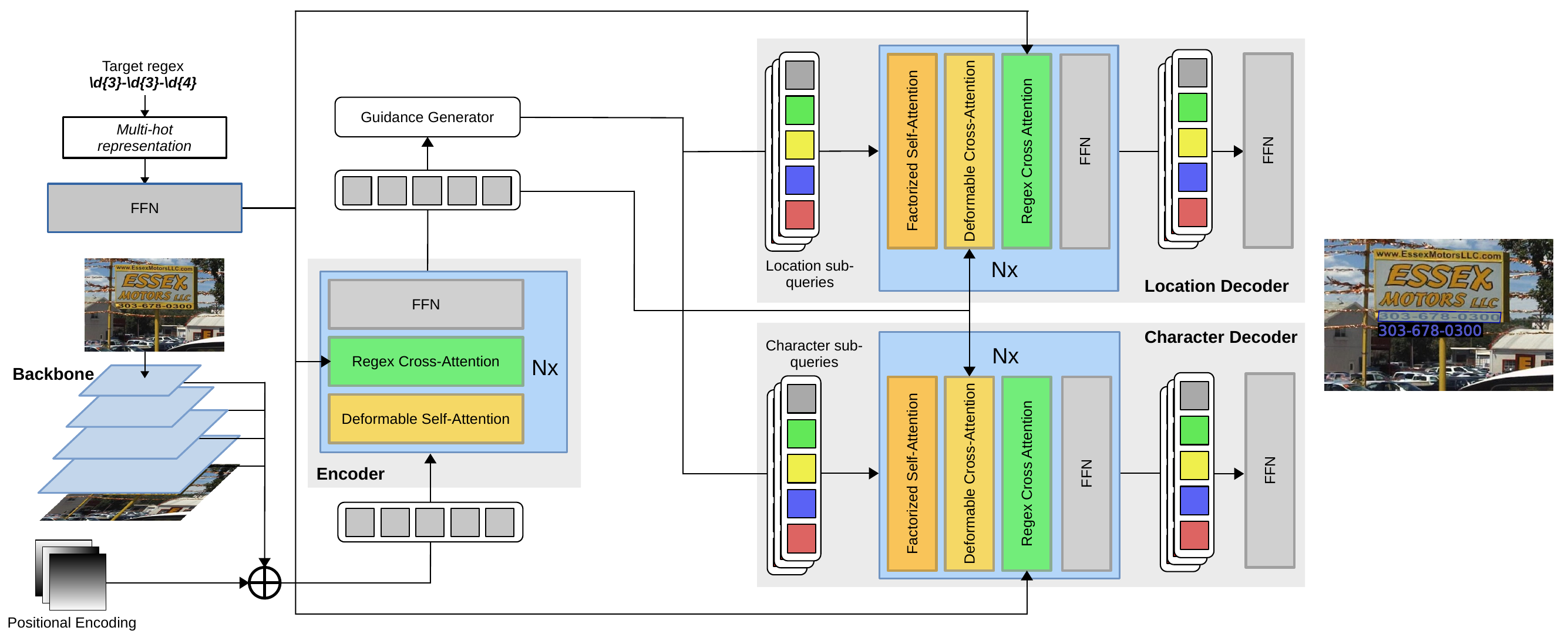}
    \caption{Detailed schematic of STEP, our proposed architecture for structured scene-text detection and recognition based on TESTR\cite{zhang2022text}.
    The features extracted by a CNN are the input to a Deformable DETR-like\cite{zhu2020deformable} encoder. 
    A cross-attention layer in the encoder combines the image features and the target structure, biasing the guidance generator to generate proposals that match the input regex.
    Two different branches perform recognition (the character decoding branch) and polygon coordinate regression (the location decoder branch) guided by cross-attention layers in the character and location decoders.}
    \label{fig:architecture}
\end{figure*}

\section{Related Work}

Our structured scene-text spotting task differs from the classic scene-text detection and recognition paradigm. Generic scene-text datasets feature annotations at arbitrary granularities (the most common being word level), while our task contains text with spaces. Scene-text architectures trained on these datasets can be applied to our task, but require post-processing operations. Since our approach is based on TESTR (a generic scene-text architecture) and trained on public data, in this section we give an overview of the state of the field.

\subsection{Scene-Text Detection and Recognition Datasets}



Most scene-text datasets feature word-level annotations. They mainly differ on the source of the images, which can be focused (such as in ICDAR13\cite{karatzas2013icdar}) or incidental text (such as in ICDAR15\cite{karatzas2015icdar}). 
One of the key differences is the type of annotations, which can include rotated quadrilaterals (found for example in ICDAR15\cite{karatzas2015icdar}, MLT 2017\cite{nayef2017icdar2017} or MLT2019\cite{nayef2019icdar2019}), or polygonal annotations (which are used on datasets that focus on irregular text such as Total-Text\cite{ch2017total} and CTW1500\cite{yuliang2017detecting}).
Other datasets like TextOCR\cite{singh2021textocr} and Open Images V5 Text\cite{krylov2021open} focus on collecting datasets with images that come from large image databases.

All the previously mentioned datasets annotate text at word level. 
HierText\cite{long2022towards}, unlike the previously mentioned datasets, contains three hierarchical levels of layout information. 
The three levels of information are paragraph, line, and word. The line-level information provides adjacency information between the words, which we have used to generate a new dataset featuring spaces. 
Section \ref{subsection:train_details} contains more information about the dataset generation and training procedure.

\subsection{Scene-Text Detection and Recognition Architectures}

Our proposed task is closely related to scene-text detection and recognition, a field that has attracted active research interest in the last few years.
The most common approach has been to use two-stage architectures to perform the detection and recognition.
TextBoxes\cite{liao2017textboxes} is an example of a two-stage architecture, where an SSD-inspired\cite{liu2016ssd} network performs detections and a CRNN\cite{shi2016end} network performs recognition. 
Since this method was limited to detecting horizontal text, FOTS\cite{liu2018fots} performs detection using multi-oriented bounding boxes. The authors introduce RoIRotate, a pooling operation that rectifies the visual features horizontally before they get recognized.
More recent architectures try to detect arbitrarily shaped scene text.
For example, TextDragon \cite{feng2019textdragon} predicts text series of quadrangles that follow the text centerline.
The ABCNet\cite{liu2020abcnet} and ABCNet v2\cite{liu2021abcnet} pipelines use a more unconventional approach by fitting Bezier curves to the text instances.

There has been a recent community trend of utilizing the powerful self-attention mechanism of the transformer\cite{vaswani2017attention} architecture.
One example of a transformer-based model is TTS\cite{kittenplon2022towards}, which uses a shared transformer encoder-decoder with different decoder heads to perform word recognition, detection, and segmentation. 
SwinTextSpotter\cite{huang2022swintextspotter} utilizes various transformer-encoder networks to enhance the interaction between detected regions and the recognizer.
Some transformer-based models, such as SPTS\cite{peng2022spts} and DEER\cite{kim2022deer}, can use basic annotations like central keypoints.
TESTR \cite{zhang2022text} also uses an encoder-decoder approach for text detection and recognition. The encoder is based on Deformable DETR\cite{zhu2020deformable} detector, while two transformer decoders perform character and polygon decoding. We have based STEP, our approach towards structured text spotting, on this architecture.

\subsection{Structured Text Spotting}


The domain gap that structured scene-text presents has been an unexplored topic in the community.
Approaches such as \cite{gomez2018cutting} can learn to recognize structured information (in this case utility meters) but require large amounts of labeled data, which is often limited or outright nonexistent.
Related to our idea of exploiting the prior knowledge of the target text, the authors of \cite{shi2021improving} opt to bias a CNN-LSTM-CTC recognizer network\cite{shi2016end} by injecting the regex of the target structure text into the model's decoder. The images are mostly documents and handwritten text. The biasing only takes place on the recognizer, while the localization comes from the line-level information of the dataset.
The authors show how this model conditioning reduces spelling mistakes.
In our work, we have focused on scene-text spotting, which requires both localization and recognition in natural images.
To the best of our knowledge, this is the first attempt at zero-shot, structure-guided scene-text spotting.

\section{Methodology}



In this section, we describe the architecture and training procedure of Scene TExt sPotter (STEP), our approach towards structured scene-text spotting. Our training and evaluation strategy uses a modified version of HierText to create our training and validation splits.

\subsection{Architecture}

The STEP architecture is based on TESTR\cite{zhang2022text}, an end-to-end framework for generic text detection and recognition. This architecture is composed of an encoder and two decoder networks based on the transformer\cite{vaswani2017attention}.
In our architecture, the encoder and decoders have been modified to make them aware of the target structure. 
This structure is queried as a regular expression, which we represent using a series of multi-hot encoded vectors.
Figure \ref{fig:architecture} shows a detailed overview of STEP.

\subsubsection{Text Format Encoding}
\label{section:regex_encoding}

One of the challenges of this task is to make the network aware of the structure of the target text. 
STEP's method of representing this structure is based on regular expressions, of which we can represent certain pattern-matching operations.
The regex representation is formulated as 
$\mathbf{H} = (\mathbf{h}_{1}, \ldots, \mathbf{h}_{M})$, 
where $M$ is the maximum recognition length and each $\mathbf{h}_{m}$ is a multi-hot encoded vector. Each vector $\mathbf{h}_{m}$ is defined as $\mathbf{h}_{m} = (h_{m, 1}, \ldots, h_{m, K})$ where $h_{m, k} \in \{0, 1\}$ and $K$ is the number of characters in our character set. An element $k$ represents a particular character of this set.
The element $h_{m,k}$ is set to $1$ if we know that our target text can have a character with index $k$ in position $m$. Since this is a multi-hot representation, we can set multiple elements of $\mathbf{h_{m}}$ to $1$.

With this multi-hot encoding we can represent certain regex operands.
One of these operations is matching the type of characters at a certain position (in regex, expressions enclosed in brackets ``[ ]''). For example, we can encode the pattern ``\texttt{\textbackslash b[A-Za-z]\{5\}}'', (any five-letter word). With the multi-hot vectors we can select combinations of different characters, allowing us to encode more general patterns like ``\texttt{[A-Za-z0-9]\{4\}}'' (any word with a combination of 4 letters or numbers) or more specific ones like ``\texttt{A\textbackslash d\{2\}0}'' (any word starting with the letter ``A'', followed by two numbers, and ending with ``0''). Likewise, we can represent the characters that we do not want to match by setting those characters to 0 (in regex, expressed with the bracket expression ``\texttt{[\string^ ]})''. For example ``\texttt{[\string^1-5]\{4\}}'', any 4-character word that does not contain numbers 1 to 5.

The representation is limited to strings of a fixed number of characters, so the operands ``\texttt{+}'' or ``\texttt{*}'' are not supported. We also can not encode the expression ``\texttt{[0-9]\{2-5\}}'', which represents any number with 2 to 5 digits. As a consequence, we can not query structures of variable length in a single forward pass.



\subsubsection{Encoder}

The TESTR encoder uses the multi-scale deformable attention module from Deformable DETR\cite{zhu2020deformable}. The multi-scale features from the CNN backbone serve as the input to this layer. The deformable attention layer only attends to a small set of keys for each query, reducing the computational complexity of the attention mechanism. The input multi-scale feature maps are defined as $\{ \mathbf{x}^{l} \}^{L}_{l=1}$, where $\mathbf{x}^{l} \in \mathbb{R}^{C \times H_{l} \times W_{l}}$ and $L$ is the level of the feature map. If $\mathbf{\hat{p}}_{q} \in [0, 1]^{2}$ are the normalized coordinates of the reference point for a query $q$, the deformable attention is defined as

\begin{equation}
    \begin{split}
        MSDeformAttn(\mathbf{z}_{q}, \mathbf{\hat{p}}_{q}, \{ \textbf{x}^{l} \}^{L}_{l=1} ) = \\
        \sum_{m=1}^{M} \mathbf{W}_{m} \left [ \sum_{l=1}^{L} \sum_{k=1}^{K} A_{mlqk} \cdot \mathbf{W}'_{m} \mathbf{x}^{l} (\phi_{l}(\mathbf{\hat{p}}_{q}) + \Delta \mathbf{p}_{mlqk}) \right ]  
    \end{split}
\end{equation}

where $m$, $l$ and $k$ are the attention head, the input feature level and the sampling point, respectively. $\mathbf{A}_{hlk}$ and $\Delta\mathbf{p}_{mlqk}$ are the attention weight and the sampling offset for query element $q$. $\phi_{l}(\mathbf{\hat{p}}_{q})$ performs a mapping from normalized image coordinates to its location in the l-th level of the feature map. $\mathbf{W}_{m}$  and $\mathbf{W}'_m$ are trainable matrices.

Like in TESTR, STEP uses the Two-Stage version of the Deformable DETR. In the Two-Scale variant, the guidance generator at the output of the encoder generates coarse bounding box proposals as the first stage. These generated proposals serve as the initialization of the object queries in the decoders, the second phase.
Each pixel of the multi-scale feature map is used to generate a proposal, but only the top-scoring bounding boxes are picked.
The generation of the bounding box proposals is supervised with an intermediate classification loss and an IoU loss.

In the regular TESTR, the bounding boxes should be generated over the areas of the image that contain text. Since the model is often trained at word granularity, it is easy for the encoder to come up with reasonably good proposals.
In our problem, however, we can not directly use this approach, since the generated boxes should be over areas of text that follow the target structure.
This requires the encoder to somehow be aware of the regex before generating the proposals.
In STEP, we have added an additional multi-head cross-attention layer in each of the encoder layers. This cross-attention uses the multi-scale image features as the queries, and the encoded regex $\mathbf{H}$ as the key and values.
This layer conditions the encoder to generate proposals over areas of text that follow the queried regex.

\subsubsection{Decoders}
\label{section:decoders}

STEP follows the idea of TESTR of tackling text detection and recognition as learning to predict tuples of points and characters. 
If $K$ is the number of proposals of the guidance generator and $i$ is the index of each proposal, the model learns to predict the tuple $Y = \{(\textbf{P}^{(i)}, \textbf{C}^{(i)}\}^{K}_{i=1}$, where $\textbf{P}^{(i)} = (p^{(i)}_{1}, \dots, p^{(i)}_{N})$ are the N coordinates of the predicted polygon and $\textbf{C}^{(i)} = (c^{(i)}_{1}, \dots, c^{(i)}_{M})$ are the M characters of the predicted text. 
The two sets of elements of the tuple are predicted by the location and character decoders, respectively.

The character decoder extends each query to $M$ sub-queries, each sub-query is a character of the recognition $\textbf{C}^{(i)} = (c^{(i)}_{1}, \dots, c^{(i)}_{M})$. The decoder is composed of a deformable cross-attention layer with the image features and factorized self-attention layers. 
The factorized self-attention, inspired by \cite{dong2021visual}, includes an intra-attention layer between the elements of $\textbf{C}^{(i)}$ and an inter-attention between the characters $c_{j}$ across different words.
A classification layer predicts the final class of each sub-query.
Additionally, STEP adds a cross-attention layer between the elements of $\textbf{C}^{(i)}$ and the embedded regex expression $\mathbf{H} = (\mathbf{h}_{1}, \ldots, \mathbf{h}_{M})$. The character sub-queries serve as the queries while vectors of the regex serve as the keys and values.
This cross-attention guides the recognition process and reduces spelling mistakes.
Similarly, the location decoder extends each instance query with $N$ sub-queries, and each sub-query is a control point of the polygon $\textbf{P}^{(i)} = (p^{(i)}_{1}, \dots, p^{(i)}_{N})$. 
Like in the character decoder, it is composed of a deformable cross-attention layer with the image features and factorized self-attention layers. 
We also add a cross-attention layer in between the sub-query points $\textbf{P}^{(i)}$ of each polygon and the vectors of the regex $\mathbf{H}$.

STEP also differs in the approach to calculating each proposal's objectness score. 
Vanilla TESTR uses each location sub-query to predict a confidence score, where the average is the final score of the instance. 
In regular scene-text detection, visual appearance may suffice to determine if a proposal overlaps with a text instance. However, in structured text spotting, the contents of the detected region are also important to ascertain the validity of a proposal. 
The text within the region must adhere to the queried structure, which means that transcription information should also be factored in. To address this, STEP utilizes both location and character sub-queries to produce the confidence score.

\subsection{Model Training}
\label{subsection:train_details}

\begin{figure}
    \centering
    \begin{tabular}{cc}
       \begin{subfigure}{0.45\columnwidth}
            \includegraphics[width=\columnwidth]{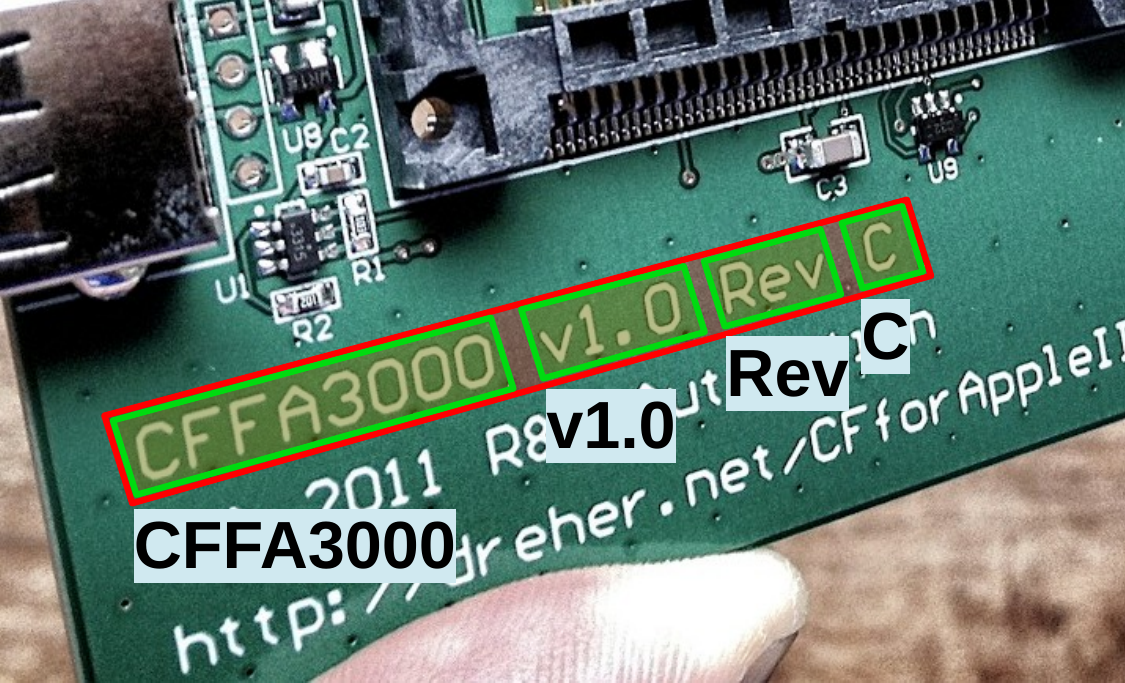}
            \caption{Hierarchical line (red) and word-level (green) annotation.}
            \label{fig:ht_custom_sub1}
       \end{subfigure}
       & 
       \begin{subfigure}{0.45\columnwidth}
            \includegraphics[width=\columnwidth]{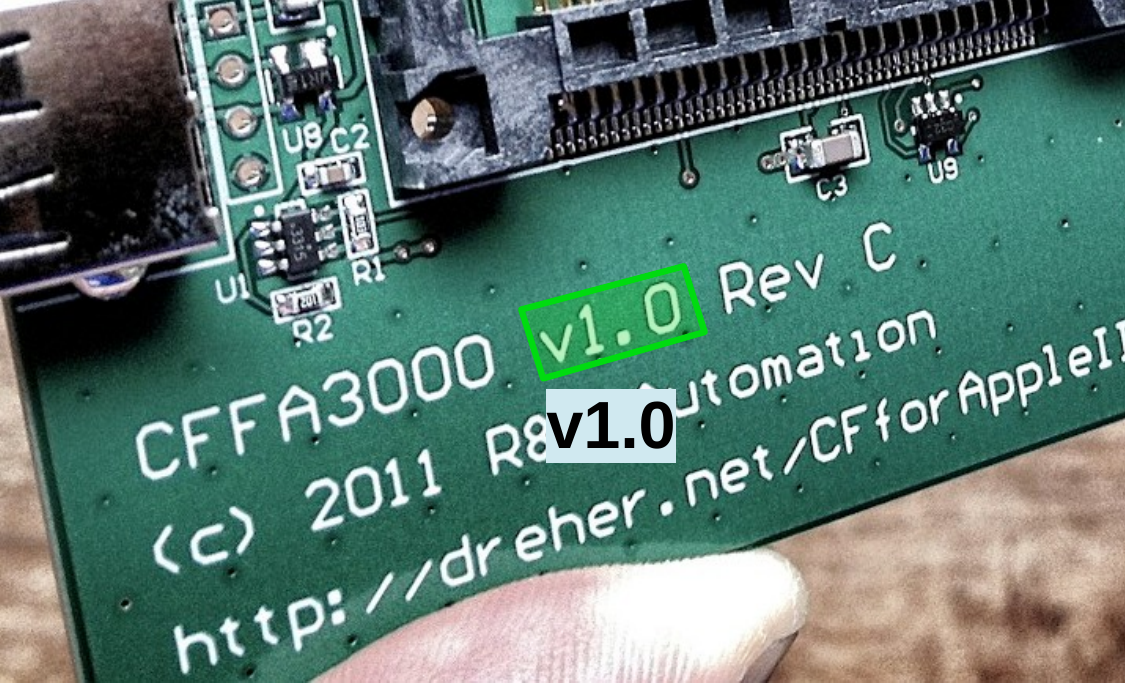}
            \caption{Selected word with at least one non-alphabetical character.}
            \label{fig:ht_custom_sub2}
       \end{subfigure}\\
       \begin{subfigure}{0.45\columnwidth}
            \includegraphics[width=\columnwidth]{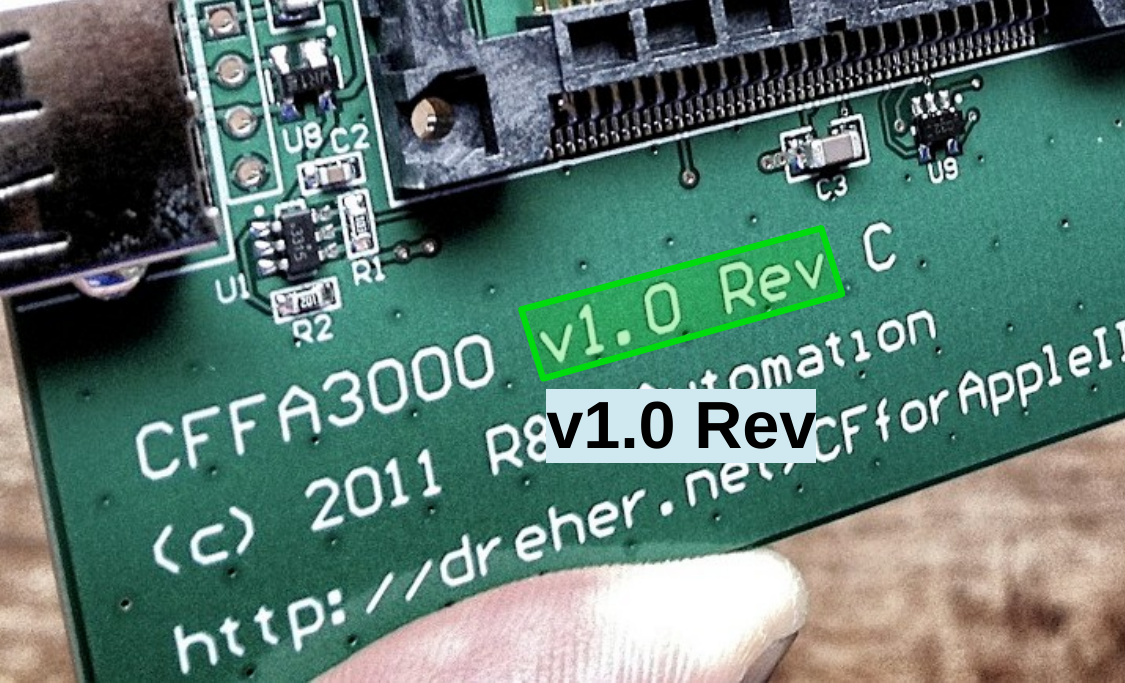}
            \caption{Merging of the selected word and its left neighbor.}
            \label{fig:ht_custom_sub3}
       \end{subfigure} &
       \begin{subfigure}{0.45\columnwidth}
            \includegraphics[width=\columnwidth]{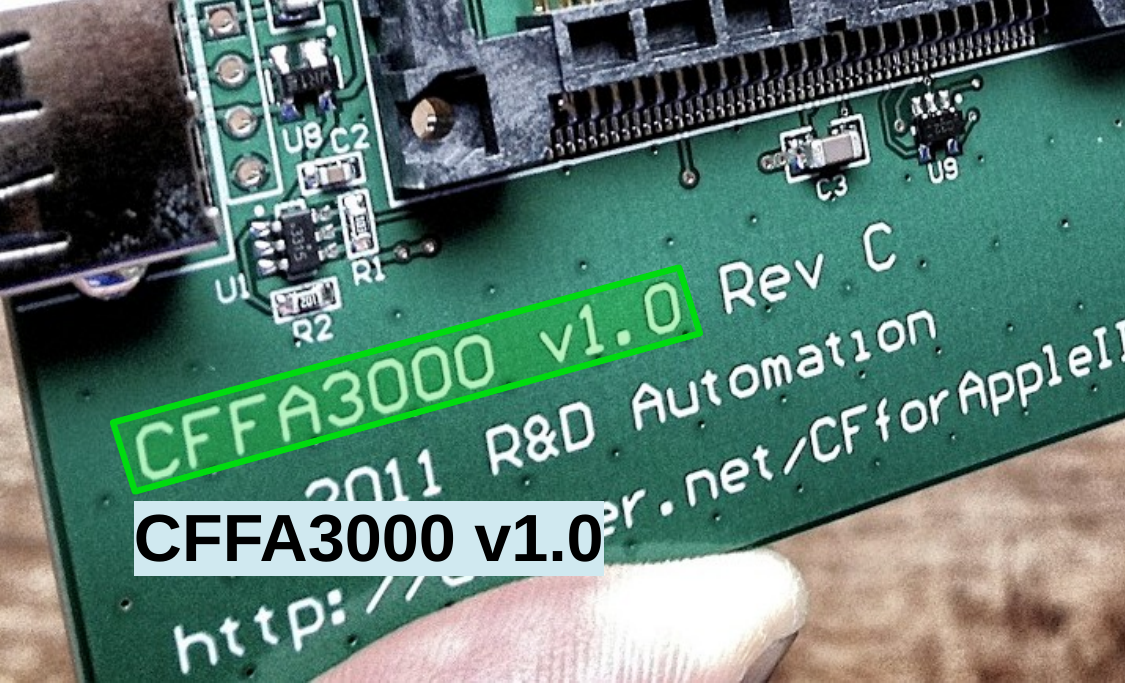}
            \caption{Merging of the selected word and its right neighbor.}
            \label{fig:ht_custom_sub4}
       \end{subfigure}
    \end{tabular}

  \caption{Our HierText-derived dataset uses line and word-level annotations to create new annotations with spaces. Starting off from a single annotated line (Figure \ref{fig:ht_custom_sub1}), we keep all the words that contain at least one non-alphabetical character (Figure \ref{fig:ht_custom_sub2}). Additionally, we also try to create new annotations by merging the selected annotation with its adjacent words. In Figures \ref{fig:ht_custom_sub3} and \ref{fig:ht_custom_sub4}, we have merged the polygons of the word ``v1.0'' with its two adjacent words. The final captions are the two sub-captions separated by a space.}
  \label{fig:ht_custom}
\end{figure}

Classic scene text datasets feature word-level annotations. Our objective is to build an OCR system that is capable of detecting and recognizing text with arbitrary structures, which can include spaces.
The HierText\cite{long2022towards} dataset is a particular case among scene-text datasets. This dataset features three levels of hierarchical annotations; paragraph, line, and word level. We have modified this dataset to create varied and challenging training and evaluation splits, which include spaces.

To reduce the bias towards in-vocabulary words (as defined by \cite{wan2020vocabulary}, words seen during the training phase) in our modified HierText dataset, we only kept the instances that contain at least one non-alphabetical character.
We also use the line-level information to merge adjacent neighbors and create new annotations with spaces.
The process includes merging the polygons and both captions, which are concatenated with a space in-between. The merging strategy is shown in Figure \ref{fig:ht_custom}.

The training pipeline of the network follows a similar strategy as other scene-text models, with the difference that our ground truth only contains the instances that match the query.
Each time the dataloader samples an image, we select one of the ground truth instances at random (the image might contain numerous), which we use to generate the regex representation $\mathbf{H}$.
For each character $m$ of the selected word, we generate its vector $\mathbf{h}_{m}$.
The elements of $\mathbf{h}_{m}$ are set to 1 depending on the type of the character (letter, number, space, etc.).
We can also randomly set just one element of $\mathbf{h}_{m}$ to 1 in order to force the specific character $m$ in that position.
Since multiple text instances can match the generated query, the rest of the words of the image are compared against the generated regex. The matching instances are included in the ground truth.
The disadvantage of this approach is that we need more training iterations in order to see all the ground truth instances of the training set.

\subsubsection{Training Details}

We use the TESTR pre-trained weights provided by the authors to initialize STEP, all the layers that are not present on the original TESTR are randomly initialized. These pre-trained weights used a mixture of SynthText 150k (coming from \cite{liu2020abcnet}), MLT 17\cite{nayef2017icdar2017} and Total-Text\cite{ch2017total}. The initial learning rate of the network is $10^{-4}$, and is decayed by a factor of 0.1 at 60k and 200k steps, the model is trained for a total of 300k steps. Like in TESTR, the learning rates of the backbone and linear projections of the reference points and sampling offsets are scaled by a factor of $0.1$. The optimizer is AdamW with $\beta_{1}=0.9$, $\beta_{2}=0.999$ and weight decay of $10^{-4}$. We used a batch size of $6$ images and the training takes around 2 days in two RTX 6000 Ada GPUs.

\subsection{Model Evaluation}
\label{section:model_evaluation}

We use the modified validation split of HierText to evaluate both our model and the generic baselines.
Like in the training split, the evaluation contains spaced text.
The evaluation protocol provides the regex of the target text to each method, which has to locate and transcribe all the matching instances. This format contains the type of each character (letter, number, space, etc.) and its length.
For example, for the string target ``Abcd 123-1'', the provided regex is ``\texttt{[a-zA-Z]\{4\}\textbackslash s\textbackslash d\{3\}-\textbackslash d}''.
Methods can make use of this information to guide the network or perform post-processing operations.
We used the classical precision and recall metrics used in object detection to evaluate on this split. 
Instances must have an intersection over union of over 0.5 with a ground truth instance to be considered a localization match. When the model is being evaluated End-To-End, the transcription of the proposal and the ground truth instance must be the same.

\section{Structured Scene-Text Dataset}
\label{section:test_dataset}

To better display the zero-shot capabilities of our approach, we introduce a new scene-text test dataset that puts the focus on out-of-vocabulary structured scene-text.
Our dataset includes $836$ images where 6 types of formatted text have been annotated. 
Much of the text contains multiple spaces and does not follow text found in any vocabulary (as opposed to generic scene-text datasets).
Since the format of each code is known, methods can make use of this prior information to condition the network or use it in post-processing operations (just like in our evaluation protocol).
Figure \ref{figure:qual_ex} shows qualitative examples of some of the formats included in this split. Section F of the supplementary material contains further information about the images and text featured in the test split.

With the exception of license plates and phone numbers, all images have been collected by us.
The images with license plates come from the UFPR-ALPR dataset\cite{laroca2018robust}. This dataset contains 150 sequences of 30 images, each one of the sequences features a vehicle with a license plate. The 150 sequences are divided into test, train, and validation. In our test dataset we have kept the first image of every sequence of the three splits, and we have discarded all the vertical license plates, ending up with a total of 121 images.
The images with phone numbers come from the Uber-text\cite{zhang2017uber} dataset, a large-scale scene-text dataset sourced from the Bing Maps Streetside program, where each text instance is labeled with a category (such as ``license plate'' or ``street number''). We have collected a total of 109 images that contain at least one phone number.


\begin{table*}[]
\centering
    \begin{tabular}{cccccccc}
        \Xhline{3\arrayrulewidth}

        \multicolumn{1}{c|}{\multirow{2}{*}{Model}} & \multicolumn{4}{c|}{End-To-End} & \multicolumn{3}{c}{Detection} \\ \cline{2-8} 
        \multicolumn{1}{c|}{} & \multicolumn{1}{l}{Precision} & \multicolumn{1}{c}{Recall} & \multicolumn{1}{c}{F-score} & \multicolumn{1}{c|}{Avg. ED} & \multicolumn{1}{c}{Precision} & \multicolumn{1}{c}{Recall} & \multicolumn{1}{c}{F-score} \\
        \Xhline{3\arrayrulewidth}
        
        \multicolumn{1}{c|}{ABCnet v2\cite{liu2021abcnet}} & \multicolumn{1}{c|}{0.72} & \multicolumn{1}{c|}{0.31} & \multicolumn{1}{c|}{0.43} & \multicolumn{1}{c|}{0.26} & \multicolumn{1}{c|}{\textbf{0.9}} & \multicolumn{1}{c|}{0.27} & \multicolumn{1}{c}{0.42} \\ 
        
        \multicolumn{1}{c|}{SwinTS\cite{huang2022swintextspotter}} & \multicolumn{1}{c|}{0.67} & \multicolumn{1}{c|}{0.27} & \multicolumn{1}{c|}{0.39} & \multicolumn{1}{c|}{0.22} & \multicolumn{1}{c|}{0.80} & \multicolumn{1}{c|}{0.32} & \multicolumn{1}{c}{0.46} \\ 
        
        \multicolumn{1}{c|}{TESTR\cite{zhang2022text}} & \multicolumn{1}{c|}{0.72} & \multicolumn{1}{c|}{0.50} & \multicolumn{1}{c|}{0.59} & \multicolumn{1}{c|}{0.19} & \multicolumn{1}{c|}{0.87} & \multicolumn{1}{c|}{0.51} & \multicolumn{1}{c}{0.64} \\ 
        
        \multicolumn{1}{c|}{\textbf{STEP}} & \multicolumn{1}{c|}{\textbf{0.78}} & \multicolumn{1}{c|}{\textbf{0.64}} & \multicolumn{1}{c|}{\textbf{0.71}} & \multicolumn{1}{c|}{\textbf{0.13}} & \multicolumn{1}{c|}{0.86} & \multicolumn{1}{c|}{\textbf{0.69}} & \multicolumn{1}{c}{\textbf{0.76}} \\
        \Xhline{3\arrayrulewidth}

    \end{tabular}
    \caption{End-To-End and Detection results on our HierText-based evaluation dataset. The column labeled ``Avg. ED'' shows the average edit distance between the recognition and the ground truth transcriptions.}
    \label{table:eval_results}
\end{table*}

\section{Experiments}

We compare our approach with TESTR\cite{zhang2022text}, SwinTextSpotter\cite{huang2022swintextspotter} and ABCNet v2\cite{liu2021abcnet}, three generic state-of-art scene-text models we use as baselines.
All these models have been fine-tuned on the vanilla HierText dataset until convergence. In section B of the supplementary material we provide the training details for each one of the baselines.
In order to deal with spaced text (which is not present in vanilla HierText), we applied post-processing operations on the detected areas of the image. 

\subsection{HierText Evaluation Dataset}

\begin{table*}
\centering
    \begin{tabular}{c|c|c|c|c|c|c|c|c}
    \Xhline{3\arrayrulewidth}
    Model & Post-Processing & BIC & UIC & TARE & Phone Num. & Tonnage & License Plate & Avg. ED\\
    \Xhline{3\arrayrulewidth}
    ABCnet v2\cite{liu2021abcnet} & \xmark & 0.01 & 0.03 & 0.41 & 0.25 & 0.34 & 0.22 & 2.03 \\
    ABCnet v2\cite{liu2021abcnet} & \cmark & 0.15 & 0.12 & 0.47 & 0.32 & 0.33 & 0.33 & 1.87 \\
    SwintTS\cite{huang2022swintextspotter} & \xmark & 0.0 & 0.02 & 0.49 & 0.4 & 0.37 & 0.24 & 1.63 \\
    SwintTS\cite{huang2022swintextspotter} & \cmark & 0.36 & 0.12 & 0.59 & 0.45 & 0.38 & 0.38 & 1.30 \\
    TESTR\cite{zhang2022text} & \xmark & 0.03 & 0.07 & 0.43 & 0.58 & 0.40 & 0.18 & 0.46 \\
    TESTR\cite{zhang2022text} & \cmark & 0.43 & \textbf{0.26} & 0.62 & 0.65 & 0.39 & 0.29 & 0.70 \\
    \textbf{STEP} & - & \textbf{0.75} & 0.24 & \textbf{0.86} & \textbf{0.68} & \textbf{0.72} & \textbf{0.55} & \textbf{0.25} \\
    \Xhline{3\arrayrulewidth}
    
    \end{tabular}
    \caption{End-to-end results on the test split. Each cell of the table shows the final F-score of each method for a particular code. The last column shows the average edit distance between the recognition and the ground truth transcriptions.}
    \label{table:id_e2e}
\end{table*}

\begin{table*}
\centering
    \begin{tabular}{c|c|c|c|c|c|c|c}
    \Xhline{3\arrayrulewidth}
    Model & Post-Processing & BIC & UIC & TARE & Phone Num. & Tonnage & License Plate \\
    \Xhline{3\arrayrulewidth}
    ABCnet v2\cite{liu2021abcnet} & \xmark & 0.03 & 0.08 & 0.46 & 0.54 & 0.35 & 0.37 \\
    ABCnet v2\cite{liu2021abcnet} & \cmark & 0.23 & 0.19 & 0.61 & 0.60 & 0.35 & 0.56 \\
    SwintTS\cite{huang2022swintextspotter} & \xmark & 0.01 & 0.10 & 0.56 & 0.65 & 0.37 & 0.46 \\
    SwintTS\cite{huang2022swintextspotter} & \cmark & 0.42 & 0.2 & 0.67 & 0.73 & 0.38 & 0.65 \\
    TESTR\cite{zhang2022text} & \xmark & 0.04 & 0.08 & 0.43 & 0.67 & 0.41 & 0.21 \\
    TESTR\cite{zhang2022text} & \cmark & 0.47 & 0.27 & 0.64 & 0.72 & 0.40 & 0.37 \\
    \textbf{STEP} & - & \textbf{0.9} & \textbf{0.71} & \textbf{0.94} & \textbf{0.83} & \textbf{0.74} & \textbf{0.79} \\
    \Xhline{3\arrayrulewidth}
    \end{tabular}
    \caption{Detection results on the test split. Each cell of the table shows the final F-score of each method for a particular code.}
    \label{table:id_det}
\end{table*}

Following the protocol described in section \ref{section:model_evaluation}, we evaluated the baselines and our structure-guided architecture.
We use the provided regex to guide the detection and recognition of our model. On the generic baselines, we use this information to perform post-processing operations on the detected text.
These post-processing operations include merging instances and filtering non-matching text. This instance merging tries to join detections in case the queried regex features a space.
Section A of the supplementary provides more details about the merging procedure.


Table \ref{table:eval_results} shows the End-To-End and Detection validation results for both the baselines and our method.
On End-To-End, our model displays higher precision than the baselines and much higher recall (12\% more than the TESTR baseline), which results in a higher F-score.
Biasing the model with the structure of the target also reduces the number of spelling mistakes, as shown in the average edit distance.
Our model also obtains better detection results than the baselines. One major disadvantage of the generic methods is that since we are using the detection's recognition to filter out irrelevant text, some spelling mistakes can harm the detection performance. Mistaking a number for a letter can make the filtering process discard a valid detection, given that it does not follow the provided structure.

\subsection{Structured Scene-Text Dataset}

This section presents the results on the structured text dataset introduced in section \ref{section:test_dataset}. Following the same approach as in the evaluation set, we provide the format of the target text for the different methods tested. As in the evaluation split, the detections of the baselines are merged if the queried regex contains one or more spaces.

Tables \ref{table:id_e2e} and \ref{table:id_det} show the End-To-End and Detection results on the test dataset. Each cell showcases the F-score for each method and code type. We include the baselines with and without post-processing operations.
When the baselines do not use post-processing operations, the results are considerably worse on formats with spaces such as the BIC and UIC codes.
Our method obtains considerably better scores in both tasks for all the code types except in end-to-end UIC codes, probably due to the difficulty to recognize long sequences (although still obtains higher results in the detection task). 

\subsection{Qualitative Examples}

\begin{figure*}[]
    \begin{tabular}{@{}l@{}@{}c@{}@{}c@{}@{}c@{}@{}c@{}@{}c@{}}
       & BIC & License Plate & Phone Number & UIC & TARE \\
       \centered{TESTR} &
       \centered{\includegraphics[height=1.30cm]{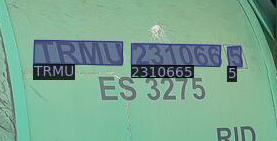}}  & 
       \centered{\includegraphics[height=1.30cm]{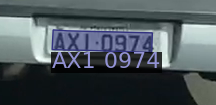}} &
       \centered{\includegraphics[height=1.30cm]{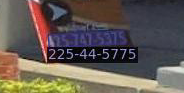}} &
       \centered{\includegraphics[height=1.30cm]{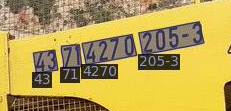}} &
       \centered{\includegraphics[height=1.30cm]{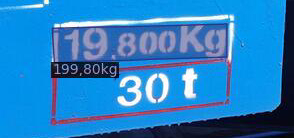}} \\

       \centered{STEP} &
       \centered{\includegraphics[height=1.30cm]{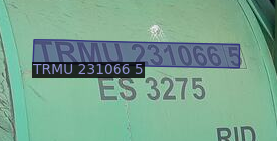}}  & 
       \centered{\includegraphics[height=1.30cm]{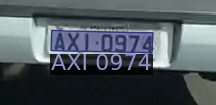}} &
       \centered{\includegraphics[height=1.30cm]{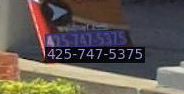}} &
       \centered{\includegraphics[height=1.30cm]{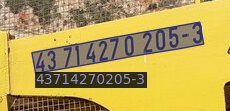}} &
       \centered{\includegraphics[height=1.30cm]{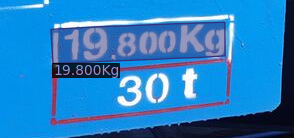}}\\
    \end{tabular}
    
    \caption{Qualitative results of TESTR and STEP on some examples from the test dataset.}
    \label{figure:qual_ex}
\end{figure*}

Figure \ref{figure:qual_ex} shows qualitative examples of TESTR and STEP on different structured codes of our test dataset. 
The TESTR results are shown without any post-processing operations except for removing irrelevant text.
The BIC and UIC codes get fragmented into different detections by TESTR, something that is not an issue with STEP. Numerous text fragmentations increase the probability of missing a part of the code, so removing this problem helps increase the recall of our method. Using the format of the code also allows our network to make fewer spelling mistakes, especially those that are related to the structure of the text. In the TARE and phone number examples, TESTR has read the wrong number of digits. Our model has less chance of getting the structure of the text wrong thanks to the cross-attention layers in the character decoder.
We also avoid confusing letters with numbers as shown in the license plate example, where TESTR has mixed the letter ``I'' for the number ``1''.

\subsection{Ablation Studies}

In this section we ablate the main architectural changes performed to TESTR. The studies have been conducted on the Hier-Text-based validation dataset.

\subsubsection{Model Confidence}

In section \ref{section:decoders} we describe how STEP uses the location and character sub-queries of an instance to calculate its classification score.
In Table \ref{table:ablation_confidence} we compare the impact on the model performance of using the character and location sub-queries. When the model uses only the character sub-queries, the F-score improves 6\% points with respect to the location-only baseline. When we use both the score is slightly improved by a 1\%.
The final version of STEP uses both modalities to calculate the detection confidence.

\begin{table}
\centering
    \begin{tabular}{c|c|c|c|c|c}
    \Xhline{3\arrayrulewidth}
    Character & Location & P & R & Fs & ED\\
    \Xhline{3\arrayrulewidth}
    \xmark & \cmark & 0.73 & 0.52 & 0.61 & \textbf{0.09} \\
     \cmark & \xmark & \textbf{0.78} & 0.60 & 0.67 & 0.11 \\ 
    \cmark & \cmark & \textbf{0.78} & \textbf{0.64} & \textbf{0.71} & 0.13 \\
    \Xhline{3\arrayrulewidth}
    \end{tabular}
    \caption{Results of using the character and location sub-queries on the HierText-based evaluation dataset.}
    \label{table:ablation_confidence}
\end{table}

\subsubsection{Regex Cross-Attention}

STEP features three cross-attention layers with the regex, one on the encoder and one in each of the two decoders. The purpose of these layers is to bias the guidance generator, the location decoder, and the character decoder. In this ablation experiment, we show the impact of removing or adding these layers to the encoder and two decoders of the network. Results are presented in Table \ref{table:ablation_cross-att}. 
As seen in the results, the encoder cross-attention is necessary to generate quality proposals. Without this layer, the guidance generator and the rest of the network are completely unaware of the structure.
Adding the cross-attention in the character decoder reduces spelling mistakes and increases the F-score by 4\%.
Finally, adding the cross-attention layer in the location decoder helps boost the F-score by 3 additional points.

\begin{table}
\centering
    \begin{tabular}{c|c|c|c|c|c|c}
    \Xhline{3\arrayrulewidth}
    Enc. & Char. & Loc. & P & R & Fs & ED\\
    \Xhline{3\arrayrulewidth}
    \xmark & \xmark & \xmark & \textbf{0.87} & 0.12 & 0.21 & \textbf{0.05} \\
    \cmark & \xmark & \xmark & 0.76 & 0.56 & 0.64 & 0.08 \\
    \cmark & \cmark & \xmark & 0.77 & 0.61 & 0.68 & 0.15 \\ 
    \cmark & \cmark & \cmark & 0.78 & \textbf{0.64} & \textbf{0.71} & 0.13 \\
    \Xhline{3\arrayrulewidth}
    \end{tabular}
    \caption{Impact of including the regex cross-attention layers on the encoder and decoders of the network. The results reported are End-To-End on the HierText-based validation set.}
    \label{table:ablation_cross-att}
\end{table}

\section{Discussion}

\subsection{Limitations}

We have shown that our approach to structured scene-text can effectively locate text with known formats in a zero-shot manner.
However, our model has some limitations, and could still be further improved beyond quantitative metrics.
In the first place, our architecture is not capable of dealing with more than one structure query at a time. Multiple queries require to do multiple forward passes. This is not a limitation of generic OCR systems, since they always produce the same readings for a given image.
Our representation of the regex is also limited to strings of a fixed length, so we can not use regex operators such as ``\texttt{+}'' or ``\texttt{*}''. 

\subsection{Conclusions}

In this paper, we have proposed the task of structured scene-text spotting, a novel structured-text test dataset, and STEP, our approach to tackle this problem. We have shown how by providing the structure to our model we can successfully guide the text-spotting process. 
The proposed method effectively removes detection fragmentations and reduces spelling mistakes.
With the training strategy proposed, the network can entirely be trained on public data and generalize well on unseen data.
This approach has been shown to be superior to using generic scene-text detection and recognition systems coupled with post-processing operations.

\paragraph{Acknowledgements} 

This work has been supported by grants PDC2021-121512-I00, PID2020-116298GB-I00 and PLEC2021-007850 funded by the European Union NextGenerationEU/PRTR and MCIN/AEI/10.13039/501100011033; the EU Lighthouse on Safe and Secure AI - ELSA funded by European Union's Horizon Europe programme under grant agreement No 101070617; the Spanish Projects NEOTEC SNEO-20211172 from CDTI and CREATEC-CV IMCBTA/2020/46; grant Torres Quevedo PTQ2019-010662; and the Industrial Doctorate programme of the Catalan Government (2020 DI 058).

{\small
\bibliographystyle{ieee_fullname}
\bibliography{egbib}
}

\end{document}


\appendix

\section{Baseline Post-Processing Operations}

In order to be able to compare the generic baselines with STEP, we had to couple them with post-processing operations. The biggest challenge was regarding queries with spaces, since generic models are usually trained at word-granularity level. The post-processing operations performed during validation and test differ from each other due to the nature of the queries and text found in them.

\subsection{Validation Split}

Our validation set contains queries with either 1 space or no spaces. When the query contains no space, all instances that do not match the queried regular expression get removed. The rest of the instances are considered positive instances.

\begin{figure*}
    \centering
    \begin{tabular}{cc}
       \begin{subfigure}{0.4\textwidth}
            \includegraphics[width=\textwidth, trim={0 0.75cm 0 0.75cm}, clip]{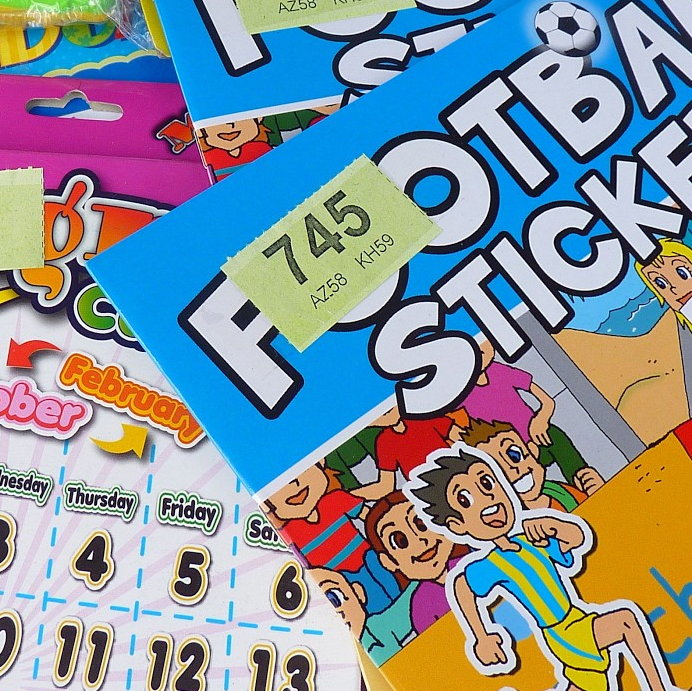}
            \caption{}
            \label{fig:post_processing_eval_1}
       \end{subfigure}
       & 
       \begin{subfigure}{0.4\textwidth}
            \includegraphics[width=\textwidth, trim={0 0.75cm 0 0.75cm}, clip]{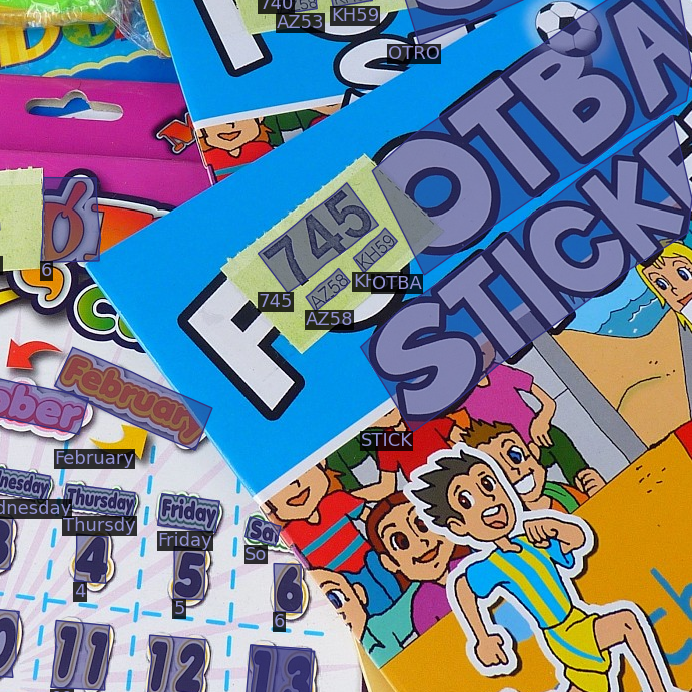}
            \caption{}
            \label{fig:post_processing_eval_2}
       \end{subfigure}\\
       \begin{subfigure}{0.4\textwidth}
            \includegraphics[width=\textwidth, trim={0 0.75cm 0 0.75cm}, clip]{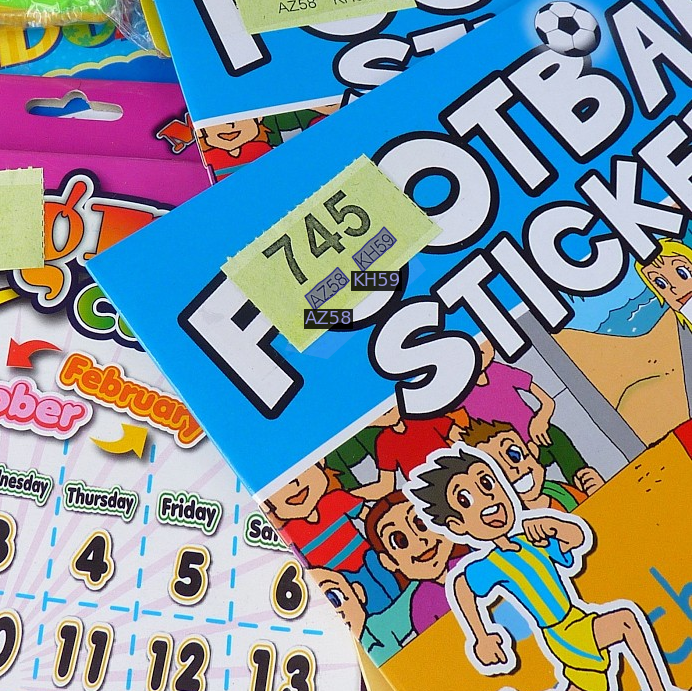}
            \caption{}
            \label{fig:post_processing_eval_3}
       \end{subfigure} &
       \begin{subfigure}{0.4\textwidth}
            \includegraphics[width=\textwidth, trim={0 0.75cm 0 0.75cm}, clip]{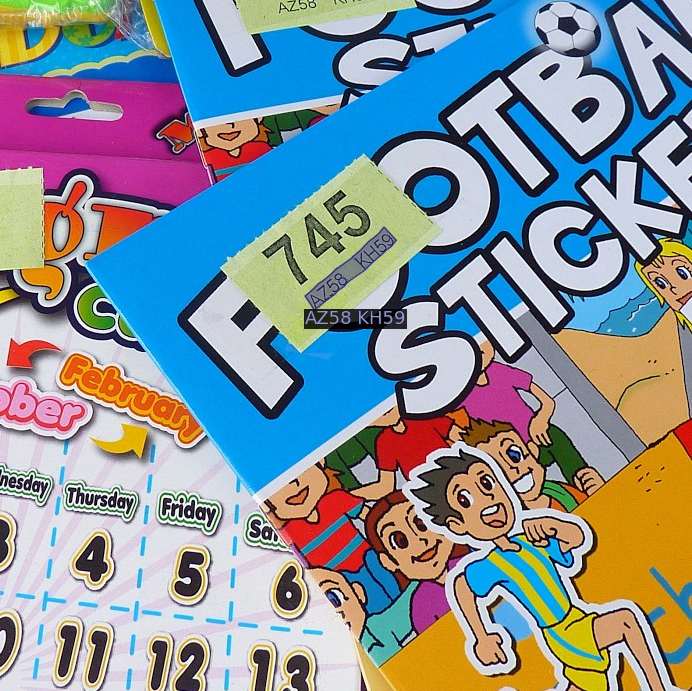}
            \caption{}
            \label{fig:post_processing_eval_4}
       \end{subfigure}
    \end{tabular}

    \caption{The target regular expression we want to find in Figure \ref{fig:post_processing_eval_1} is ``[A-Za-z]\{2\}\textbackslash d\{2\}[ ][A-Za-z]\{2\}\textbackslash d\{2\}''. Figure \ref{fig:post_processing_eval_2} shows the raw output of TESTR. The output instances are filtered by matching the two sub-queries resulting in splitting the main query by the space (in this case, ``[A-Za-z]\{2\}\textbackslash d\{2\}'' for both sub-queries), shown in figure \ref{fig:post_processing_eval_3}. The two remaining instances are merged since their distance is lower than the established threshold. Figure \ref{fig:post_processing_eval_4} shows the final polygons and transcriptions merged into a single instance.}
    \label{fig:post_processing_eval}
\end{figure*}

When a query contains a space we split it into two sub-queries. Each sub-query is one side of the query, the space being the splitting point. The output instances of the model are filtered out if they do not match one of the two sub-queries. Once the instances are filtered, we merge pair of instances that match the sub-queries and are closer than a certain threshold. We merge both the polygons of the instances and the translation with a space between transcriptions. Figure \ref{fig:post_processing_eval} shows how these post-processing operations are performed on an example from the validation set.

\subsection{Test Split}

The queries featured in the test split can contain more than one space, as seen in the example formats in Table \ref{table:ind_test_stats}. Furthermore, some instances feature arbitrary separations between the different characters of the code that do not follow any specific format (see examples in Figure \ref{figure:examples_test}), which can cause unpredictable detection fragmentations. See for example the TESTR's row in Figure \ref{figure:qual_appendix}, where the TARE and UIC codes have been fragmented at arbitrary points. 

\begin{figure*}
    \centering

     \begin{subfigure}{0.4\textwidth}
        \includegraphics[width=\textwidth, trim={0 0.75cm 0 0.75cm}, clip]{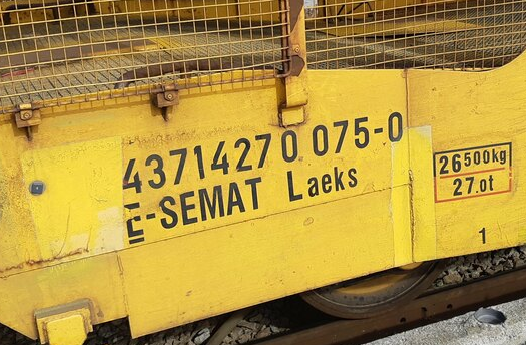}
        \caption{}
        \label{fig:post_processing_test_1}
    \end{subfigure}
    
    \begin{tabular}{cc}
      
       \begin{subfigure}{0.4\textwidth}
            \includegraphics[width=\textwidth, trim={0 0.75cm 0 0.75cm}, clip]{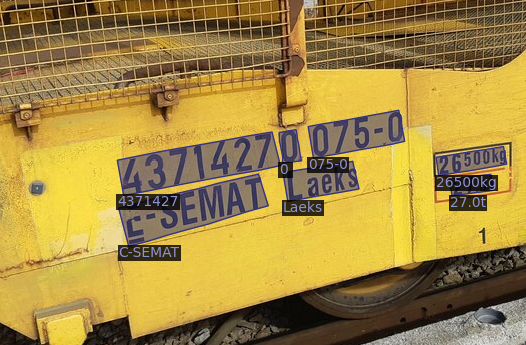}
            \caption{}
            \label{fig:post_processing_test_2}
       \end{subfigure}
       & 
       \begin{subfigure}{0.4\textwidth}
            \includegraphics[width=\textwidth, trim={0 0.75cm 0 0.75cm}, clip]{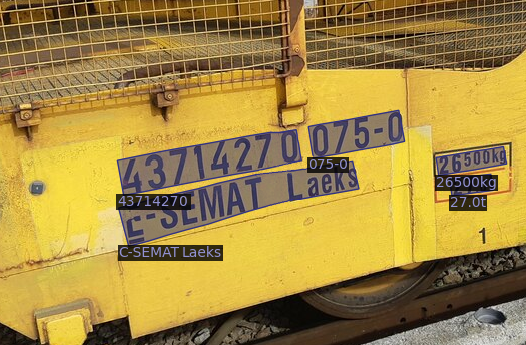}
            \caption{}
            \label{fig:post_processing_test_25}
       \end{subfigure}\\

       \begin{subfigure}{0.4\textwidth}
            \includegraphics[width=\textwidth, trim={0 0.75cm 0 0.75cm}, clip]{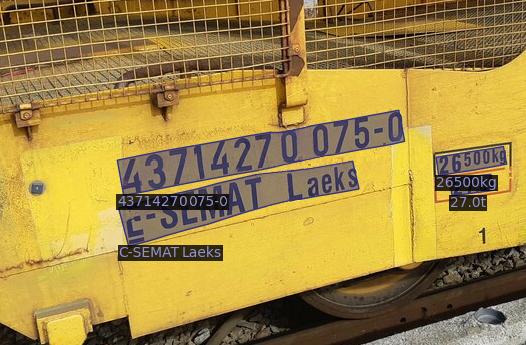}
            \caption{}
            \label{fig:post_processing_test_3}
       \end{subfigure} &
       \begin{subfigure}{0.4\textwidth}
            \includegraphics[width=\textwidth, trim={0 0.75cm 0 0.75cm}, clip]{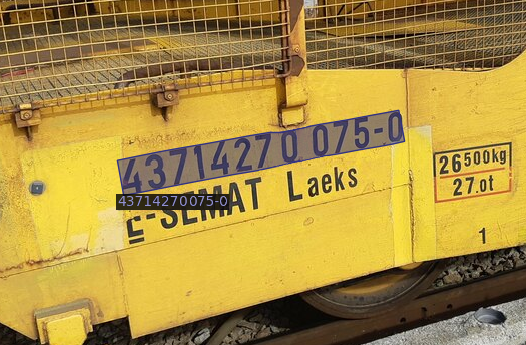}
            \caption{}
            \label{fig:post_processing_test_4}
       \end{subfigure}
    \end{tabular}

    \caption{The target regular expression we want to find in Figure \ref{fig:post_processing_test_1} is ``\textbackslash d\{11\}-\textbackslash d''. Figure \ref{fig:post_processing_test_2} shows the raw output of TESTR. Figures \ref{fig:post_processing_test_25} and \ref{fig:post_processing_test_3} show two iterations of the instance merging process. In the first Figure, it has merged part of the UIC target code (the 12-number code) that was fragmented. The next step has fully merged the remaining bit of the code. Since the resulting transcription matches the query, this instance would not be further merged with others. Other instances that do not follow the query are removed, as seen in figure \ref{fig:post_processing_test_4}.}
    \label{fig:post_processing_test}
\end{figure*}

In order to merge the detections under such circumstances, we have opted for a different strategy. First, we check if any of the unmerged instances match the query. Next, we iteratively merge pairs of instances that are close to each other (within a certain threshold). After each iteration, we check if the newly merged pairs match the query. The process ends when we can not merge any more instances. Instances that have been matched are not merged with more instances in the next iteration. This process is illustrated in Figure \ref{fig:post_processing_test}.

\section{Baseline Training Details}

In all the baselines we use pre-trained weights provided by the authors to initialize the networks and fine-tuned on the vanilla HierText training split. Unless explicitly stated, we use the default training settings used in the original implementations.

On ABCnet v2\cite{liu2020abcnet} we use the provided pre-trained weights on their SynthText 150k dataset and MLT 17. We converted the HierText polygonal annotations to the ABCnet bezier annotations. We fine-tune on HierText for 50k iterations with an initial learning rate of $10^{-3}$, which is decayed by a factor of $0.1$ at 20k and 40k iterations and a batch size of $9$ images. Since HierText features some images with dense text, we increase the maximum number of proposals per image to $300$.

For SwinTextSpotter\cite{huang2022swintextspotter} we use the provided pre-trained weights trained on MLT 17 and the SynthText 150k datasets. We fine-tune the model on HierText for 35k iterations with an initial learning rate of $10^{-4}$, decayed by $0.1$ at step 35k, and a batch size of $4$. We also increase the number of proposals per image to $300$.

Finally, we use the pre-trained weights provided by the authors of TESTR which contains, as we said, the SynthText 150k, MLT 17\cite{nayef2017icdar2017} and Total-Text datasets. The model is trained for 30k iterations with an initial learning rate of $10^{-4}$, a learning rate decay of $0.1$ at iteration 25k, and a batch size of $6$ images. The maximum number of queries is increased to $300$.

\section{Architecture Details}

STEP uses a ResNet-50\cite{he2016deep} as the feature extraction backbone of the network. We use the same setup as TESTR\cite{zhang2022text}, the encoder contains a deformable transformer\cite{zhu2020deformable} with $6$ layers, $8$ heads, and $4$ sampling points. Our encoder also contains a regular cross-attention layer with the queried regex, which also has $6$ layers and $8$ heads.
Each one of the two decoder layers is composed of a deformable cross-attention layer between the features and the queries, the intra and inter-self attention layer of the queries, and the cross-attention layer between the queries and the encoded regex. All the decoder attention blocks have $6$ layers and $8$ heads, the deformable attention also uses $4$ sampling points. The embedding dimension is $256$ in all the cases. The number of queries of the decoders is $100$.

The regex representation embedding is a feed-forward neural network that projects each one of the $\mathbf{h}_{m}$ vectors to the embedding size of the tokens. This layer is composed of $2$ hidden fully connected layers of $256$ dimensions and an output layer of the same size. Each layer has a ReLU activation function.

\subsection{Model Loss}

Our model follows the same training objectives described in \cite{zhang2022text}. The decoder training losses include, for each sub-query $j$, an instance classification loss $\mathcal{L}_{cls}^{j}$, an L1 distance loss  $\mathcal{L}_{coord}^{j}$ for control-point regression, and a cross-entropy based character classification loss $\mathcal{L}_{char}^{j}$. Opposed to TESTR, the $\mathcal{L}_{cls}^{j}$ loss uses both the character and location sub-queries to calculate the classification confidence, as opposed to location-only. The final decoder loss is $\mathcal{L}_{dec}^{j} = \sum_{j}(\lambda_{cls}\mathcal{L}_{cls}^{j} + \lambda_{coord}\mathcal{L}_{coord}^{j} + \lambda_{char}\mathcal{L}_{char}^{j})$. $\lambda_{cls}$, $\lambda_{coord}$ and $\lambda_{char}$ are used to weight the losses. We use the original values of $\lambda_{cls} = 2.0$, $\lambda_{coord} = 5.0$ and $\lambda_{char} = 4.0$. The proposals of the guidance generator of the encoder are also supervised with an instance classification loss $\mathcal{L}_{cls}^{j}$ and a coordinate regression loss $\mathcal{L}_{coord}^{j}$. Additionally, the encoder loss uses the generalized IoU loss $\mathcal{L}_{gIoU}^{j}$ defined by \cite{rezatofighi2019generalized} for bounding box regression. The encoder loss is defined as  $\mathcal{L}_{enc}^{i} = \sum_{j}(\lambda_{cls}\mathcal{L}_{cls}^{i} + \lambda_{coord}\mathcal{L}_{coord}^{i} + \lambda_{gIoU}\mathcal{L}_{gIoU}^{i})$, with $\lambda_{gIoU} = 2.0$.

\section{Regex encoding}

With the multi-hot encoding of the regex we can represent different matching operations. As described in section 3.1.1, the encoding $\mathbf{H}$ is composed by $M$ multi-hot encoded vectors $\mathbf{H} = (\mathbf{h}_{1}, \ldots, \mathbf{h}_{M})$. Each vector $\mathbf{h}_{m}$ has a total of $K$ elements $\mathbf{h}_{m} = (h_{m, 1}, \ldots, h_{m, K})$, where $K$ is equal to the size of our character set. Each element $k$ corresponds to a character of the set, and an element $h_{m, k}$ is set to 1 if that character matches the queried regex in that position. This representation provides a very fine-grained level of information at each position $m$ but, while a flexible and powerful representation, it might be difficult to learn.

As an alternative to this approach, we have also tried a less complex encoding of the regex. Instead of using the multi-hot representation, we have tried a simpler and smaller one-hot approach.
We redefine each vector $\mathbf{h}_{m}$ as $\mathbf{h}_{m} = (h_{m, 1}, \ldots, h_{m, C})$, where $C$ is equal to the number of classes we can represent at each character position $m$. We have defined a total of $6$ classes; space, number, letter, separator (characters \texttt{, -} and \texttt{\_}), special (rest of the alphabet), and padding. The disadvantage of this representation is that we no longer can match specific characters at a certain position. For example, the query ``\texttt{A\textbackslash d\{2\}0}'' would not be possible.

Table \ref{table:eval_one_vs_multi} shows the End-To-End and Detection results on the evaluation set. The multi-hot approach obtains better results in both the Detection and End-To-End tasks, although the average edit distance is slightly lower using the one-hot encoding.

\begin{table*}[]
    \centering
    \begin{tabular}{cccccccc}
        \Xhline{3\arrayrulewidth}

        \multicolumn{1}{c|}{\multirow{2}{*}{Encoding}} & \multicolumn{4}{c|}{End-To-End} & \multicolumn{3}{c}{Detection} \\ \cline{2-8} 
        \multicolumn{1}{c|}{} & \multicolumn{1}{c}{Precision} & \multicolumn{1}{c}{Recall} & \multicolumn{1}{c}{F-score} & \multicolumn{1}{c|}{Avg. ED} & \multicolumn{1}{c}{Precision} & \multicolumn{1}{c}{Recall} & \multicolumn{1}{c}{F-score} \\
        \Xhline{3\arrayrulewidth}
        \multicolumn{1}{c|}{One-Hot} & \multicolumn{1}{c|}{0.72} & \multicolumn{1}{c|}{0.57} & \multicolumn{1}{c|}{0.63} & \multicolumn{1}{c|}{\textbf{0.11}} & \multicolumn{1}{c|}{0.79} & \multicolumn{1}{c|}{0.63} & \multicolumn{1}{c}{0.73} \\ 

        \multicolumn{1}{c|}{Multi-Hot} & \multicolumn{1}{c|}{\textbf{0.78}} & \multicolumn{1}{c|}{\textbf{0.64}} & \multicolumn{1}{c|}{\textbf{0.71}} & \multicolumn{1}{c|}{0.13} & \multicolumn{1}{c|}{\textbf{0.86}} & \multicolumn{1}{c|}{\textbf{0.69}} & \multicolumn{1}{c}{\textbf{0.76}} \\
        \Xhline{3\arrayrulewidth}

    \end{tabular}
    \caption{End-To-End and Detection detections on the evaluation split using the One-Hot and Multi-Hot regex encoding approaches.}
    \label{table:eval_one_vs_multi}
\end{table*}

\section{Further Qualitative Analysis}

\subsection{Baselines Comparison}

\begin{figure*}[]
    \begin{tabular}{@{}c@{}@{}c@{}@{}c@{}@{}c@{}@{}c@{}@{}c@{}}
       & BIC & Phone Number & TARE & Tonnage & UIC \\

       \centered{ABCnet v2} &
       \centered{\includegraphics[height=1.1cm]{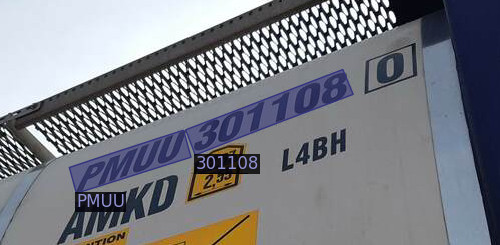}}  & 
       \centered{\includegraphics[height=1.1cm]{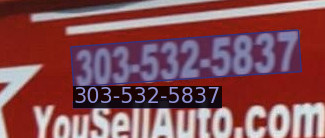}} &
       \centered{\includegraphics[height=1.1cm]{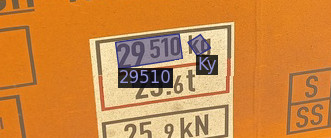}} &
       \centered{\includegraphics[height=1.1cm]{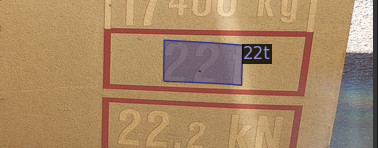}} &
       \centered{\includegraphics[height=1.1cm]{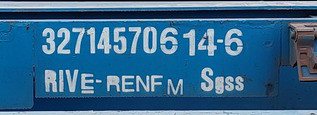}}\\
       
       \centered{SwinTS} &
       \centered{\includegraphics[height=1.1cm]{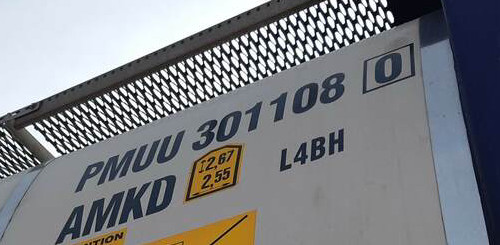}}  & 
       \centered{\includegraphics[height=1.1cm]{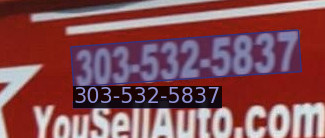}} &
       \centered{\includegraphics[height=1.1cm]{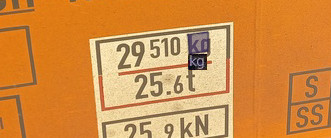}} &
       \centered{\includegraphics[height=1.1cm]{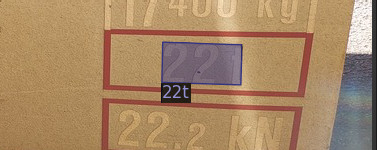}} &
       \centered{\includegraphics[height=1.1cm]{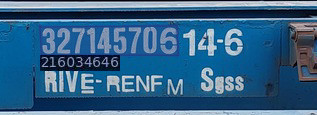}}\\

       \centered{TESTR} &
       \centered{\includegraphics[height=1.1cm]{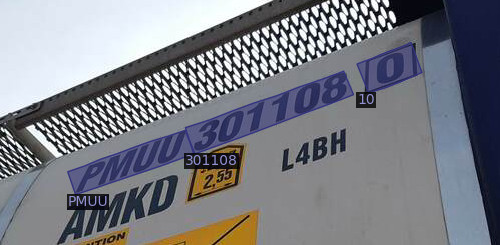}}  & 
       \centered{\includegraphics[height=1.1cm]{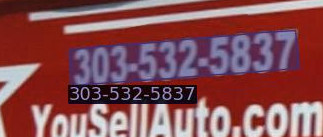}} &
       \centered{\includegraphics[height=1.1cm]{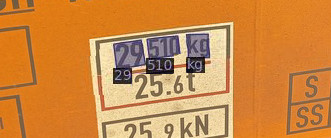}} &
       \centered{\includegraphics[height=1.1cm]{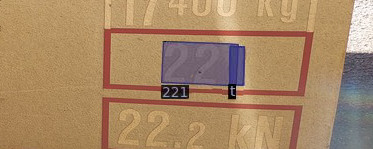}} &
       \centered{\includegraphics[height=1.1cm]{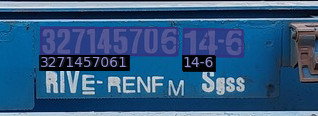}} \\

              \centered{STEP} &
       \centered{\includegraphics[height=1.1cm]{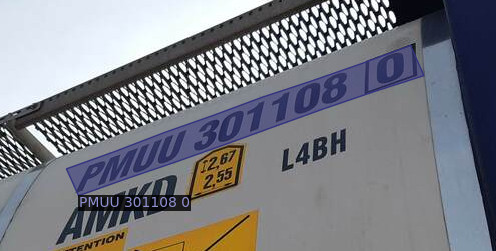}}  & 
       \centered{\includegraphics[height=1.1cm]{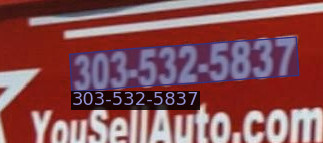}} &
       \centered{\includegraphics[height=1.1cm]{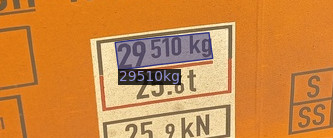}} &
       \centered{\includegraphics[height=1.1cm]{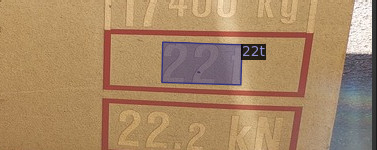}} &
       \centered{\includegraphics[height=1.1cm]{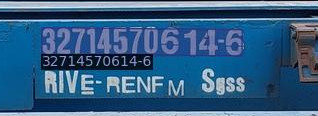}}\\
    \end{tabular}
    
    \caption{Qualitative results of the baselines and STEP on some examples from the test set.}
    \label{figure:qual_appendix}
\end{figure*}

Figure \ref{figure:qual_appendix} shows additional qualitative results of the baselines and STEP on samples from the test set. In the baseline results, we have not yet applied post-processing operations to showcase the raw output of these models. This Figure shows how the conditioning of our model results in a single detection, reducing the chance of detection failures and the need for post-processing operations.

\subsection{Failure Cases}

\begin{figure*}
    \centering
    \begin{tabular}{cccc}
        \begin{subfigure}{0.42\columnwidth}
            \includegraphics[width=\columnwidth]{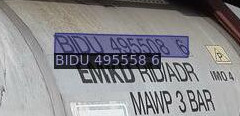}
            \caption{}
            \label{fig:qual_appendix_failures_1}
        \end{subfigure} & 

        \begin{subfigure}{0.42\columnwidth}
            \includegraphics[width=\columnwidth]{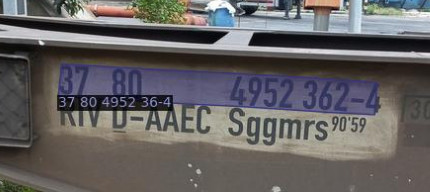}
            \caption{}
            \label{fig:qual_appendix_failures_2}
        \end{subfigure} & 

        \begin{subfigure}{0.42\columnwidth}
            \includegraphics[width=\columnwidth]{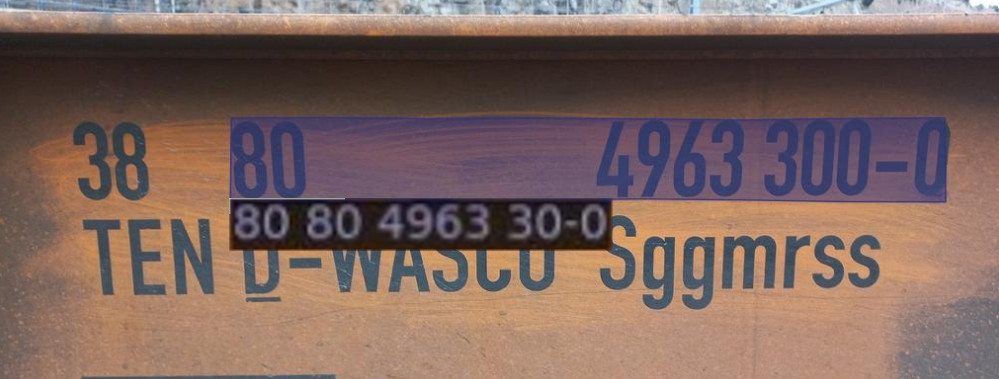}
            \caption{}
            \label{fig:qual_appendix_failures_3}
        \end{subfigure} & 

        \begin{subfigure}{0.42\columnwidth}
            \includegraphics[width=\columnwidth]{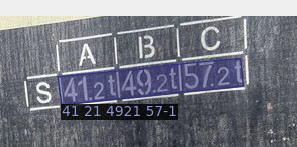}
            \caption{}
            \label{fig:qual_appendix_failures_4}
        \end{subfigure} \\
    \end{tabular}

    \caption{Qualitative examples where STEP has failed to read a BIC code (\ref{fig:qual_appendix_failures_1}), a UIC code (\ref{fig:qual_appendix_failures_2} and \ref{fig:qual_appendix_failures_3}), and produced a false positive using the UIC regex (\ref{fig:qual_appendix_failures_4}).}
    \label{fig:qual_appendix_failures}
\end{figure*}

Figure \ref{fig:qual_appendix_failures} shows typical failure cases of STEP. Figure \ref{fig:qual_appendix_failures_1} shows a misspelled BIC code. While STEP's regex-based approach reduces spelling mistakes (such as mixing certain numbers and letters), it still can generate a wrong transcription among characters of the same type. Figure \ref{fig:qual_appendix_failures_2} shows a UIC failure case where the transcription has the wrong format (one of the digits is missing). UIC codes pose a particular challenge due to their length and multiple spaces. In figure \ref{fig:qual_appendix_failures_3} the model has wrongly detected and transcribed one of such cases. STEP is also not exempt from false positives, in Figure \ref{fig:qual_appendix_failures_4} the model has been given the regex of a UIC code and wrongly read unrelated text.

\subsection{Multiple Instances}

\begin{figure*}
    \centering
    \begin{tabular}{cc}
        \begin{subfigure}{0.42\linewidth}
            \includegraphics[width=\columnwidth]{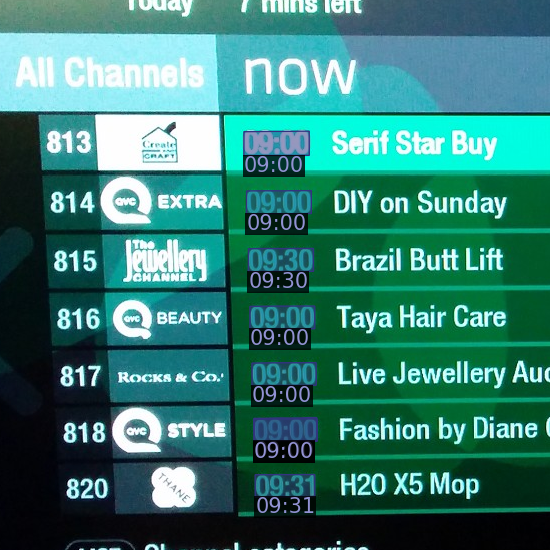}
            \caption{Query: ``\textbackslash d\{2\}:\textbackslash d\{2\}''}
            \label{fig:multi_1}
        \end{subfigure} & 

        \begin{subfigure}{0.42\linewidth}
            \includegraphics[width=\columnwidth]{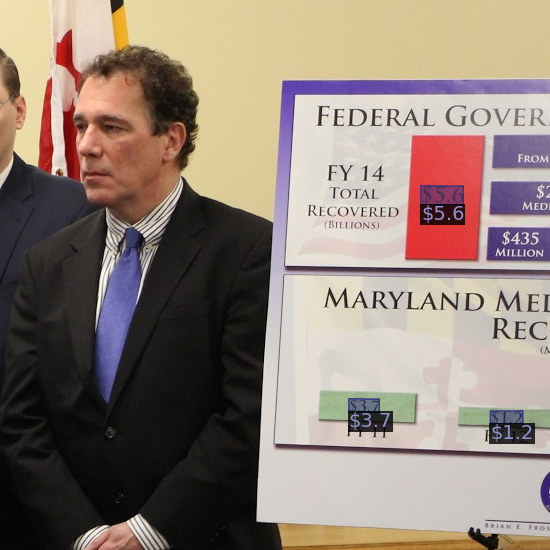}
            \caption{Query: ``\$\textbackslash d.\textbackslash d''}
            \label{fig:multi_2}
        \end{subfigure}
    \end{tabular}
    \caption{Our model is capable of detecting and reading multiple targets with a single query.}
    \label{fig:multi}
\end{figure*}

As Figure \ref{fig:multi} shows, our model is capable of successfully detecting and recognizing multiple targets for a single query. During training, we match the generated regex with all the instances of the ground truth. Since multiple instances can match this query, our model learns that a single query can have multiple instances as the ground truth.

\section{Test Dataset Details}

\begin{figure*}[]
    \centering
    \begin{tabular}{ccccc}
        BIC & Phone Number & TARE & Tonnage & UIC \\
        \includegraphics[height=1.2cm]{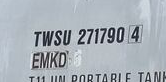}  & 
        \includegraphics[height=1.2cm]{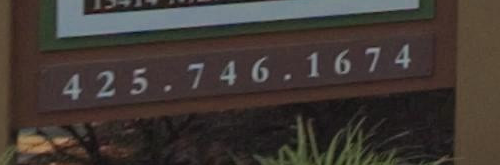} &
        \includegraphics[height=1.2cm]{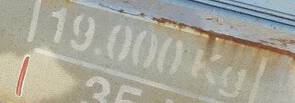} &
        \includegraphics[height=1.2cm]{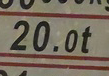} &
        \includegraphics[height=1.2cm]{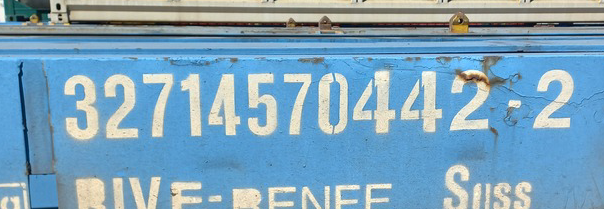}\\
       
        \includegraphics[height=1.2cm]{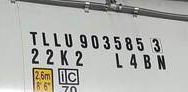}  & 
        \includegraphics[height=1.2cm]{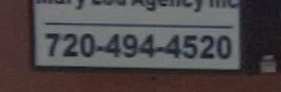} &
        \includegraphics[height=1.2cm]{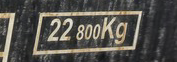} &
        \includegraphics[height=1.2cm]{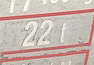} &
        \includegraphics[height=1.2cm]{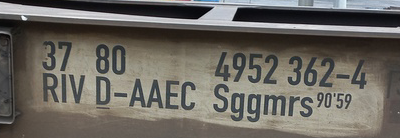}\\

        \includegraphics[height=1.2cm]{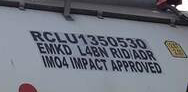}  & 
        \includegraphics[height=1.2cm]{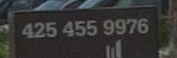} &
        \includegraphics[height=1.2cm]{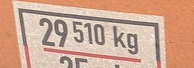} &
        \includegraphics[height=1.2cm]{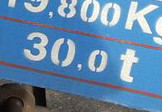} &
        \includegraphics[height=1.2cm]{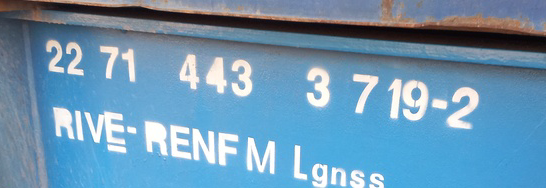} \\

    \end{tabular}
    
    \caption{Examples of the different codes featured in our test dataset. The format of some codes, such as the UIC or BIC codes, feature different separations and spaces in their format.}
    \label{figure:examples_test}
\end{figure*}

Table \ref{table:ind_test_stats} shows the number of instances for each one of the 6 codes and an example of their formats. Figure \ref{figure:examples_test} also shows more examples of the different varieties of the codes of the test set.

\begin{table}
    \begin{tabular}{c|c|c}
        \Xhline{3\arrayrulewidth}
        Code Type & Num. & Example Format  \\
        \Xhline{3\arrayrulewidth}
        BIC & 329 & ``BICU 342894 0'' \\
        UIC & 407 & ``2837 58 47 391-1''\\
        TAREs & 382 & ``25.000 KG'' or ``25000 kg'' \\
        Phone Numbers & 109 & ``123-456-7890'' \\
        Tonnage & 659 & ``25.0t'' or ''25t`` \\
        License Plates & 121 & ``ABC 1234'' \\ 
        \Xhline{3\arrayrulewidth}
    \end{tabular}
    \caption{Codes featured in our test split and examples of their format.}
    \label{table:ind_test_stats}
\end{table}

{\small
\bibliographystyle{ieee_fullname}
\bibliography{egbib}
}